\documentclass[pdflatex]{sn-jnl}

\usepackage{graphicx}
\usepackage{amsmath,amssymb,amsfonts}
\usepackage{booktabs}
\usepackage{svg}
\usepackage{multicol}
\usepackage{multirow}
\usepackage{rotating}
\usepackage{colortbl}
\usepackage{geometry}
\usepackage{xcolor}

 \newcommand{\invisible}[1]{}

\newcommand{\citewithlink}[2]{%
  \cite{#1}%
}

\usepackage{breakurl}

\makeatletter
\renewcommand{\burl}[1]{%
  \begingroup
    \def\UrlLeft{}%
    \def\UrlRight{}%
    \urlstyle{rm}
    \Urlmuskip=0mu plus 1mu
    \url{#1}%
  \endgroup
}

\g@addto@macro{\UrlBreaks}{%
  \do\a\do\b\do\c\do\d\do\e\do\f\do\g\do\h\do\i\do\j\do\k\do\l\do\m%
  \do\n\do\o\do\p\do\q\do\r\do\s\do\t\do\u\do\v\do\w\do\x\do\y\do\z%
  \do\A\do\B\do\C\do\D\do\E\do\F\do\G\do\H\do\I\do\J\do\K\do\L\do\M%
  \do\N\do\O\do\P\do\Q\do\R\do\S\do\T\do\U\do\V\do\W\do\X\do\Y\do\Z%
  \do\0\do\1\do\2\do\3\do\4\do\5\do\6\do\7\do\8\do\9\do\-}
\g@addto@macro{\UrlBreaks}{\UrlOrds}
\makeatother

\usepackage{url}

\title{Beyond Ethics: How Inclusive Innovation Drives Economic Returns in Medical AI}

\author[1]{\small{Balagopal Unnikrishnan$^{\dagger}$}}
\author[2]{\small{Ariel Guerra Adames$^{\dagger}$}}
\author[3]{\small{Amin Adibi$^{\dagger}$}}
\author[4]{\small{Sameer Peesapati}}
\author[5]{\small{Rafal Kocielnik}}
\author[6]{\small{Shira Fischer}}
\author[7]{\small{Hillary Clinton Kasimbazi}}
\author[8,9]{\small{Rodrigo Gameiro}}
\author[10]{\small{Alina Peluso}}
\author[8]{\small{Chrystinne Oliveira Fernandes}}
\author[11,8]{\small{Maximin Lange}}
\author[12,13]{\small{Lovedeep Gondara}}
\author[8,14,15]{\small{Leo Anthony Celi}}

\affil[1]{\small{Department of Computer Science, University of Toronto, Canada}}
\affil[2]{\small{Bordeaux Population Health Research Center, University of Bordeaux, France}}
\affil[3]{\small{Faculty of Pharmaceutical Sciences, University of British Columbia, Canada}}
\affil[4]{\small{Synthesize Health, Canada}}
\affil[5]{\small{Computing and Mathematical Sciences, California Institute of Technology, USA}}
\affil[6]{\small{RAND Corporation, USA}}
\affil[7]{\small{Department of Radiology and Radiotherapy, Makerere University, Uganda}}
\affil[8]{\small{Laboratory for Computational Physiology, Massachusetts Institute of Technology, USA}}
\affil[9]{\small{Department of Bioengineering, Instituto Superior Técnico, Portugal}}
\affil[10]{\small{Oak Ridge National Laboratory, USA}}
\affil[11]{\small{Institute of Psychiatry, Psychology \& Neuroscience, King's College London, UK}}
\affil[12]{\small{School of Population and Public Health, University of British Columbia, Canada}}
\affil[13]{\small{Provincial Health Services Authority, British Columbia, Canada}}
\affil[14]{\small{Division of Pulmonary, Critical Care and Sleep Medicine, Beth Israel Deaconess Medical Center, USA}}
\affil[15]{\small{Department of Biostatistics, Harvard T.H. Chan School of Public Health, USA}}
\abstract{While ethical arguments for fairness in healthcare AI are well-established, the economic and strategic value of inclusive design remains underexplored. This perspective introduces the ``inclusive innovation dividend''---the counterintuitive principle that solutions engineered for diverse, constrained use cases generate superior economic returns in broader markets. Drawing from assistive technologies that evolved into billion-dollar mainstream industries, we demonstrate how inclusive healthcare AI development creates business value beyond compliance requirements. We identify four mechanisms through which inclusive innovation drives returns: (1) market expansion via geographic scalability and trust acceleration; (2) risk mitigation through reduced remediation costs and litigation exposure; (3) performance dividends from superior generalization and reduced technical debt, and (4) competitive advantages in talent acquisition and clinical adoption. We present the Healthcare AI Inclusive Innovation Framework (HAIIF), a practical scoring system that enables organizations to evaluate AI investments based on their potential to capture these benefits. HAIIF provides structured guidance for resource allocation, transforming fairness and inclusivity from regulatory checkboxes into sources of strategic differentiation. Our findings suggest that organizations investing incrementally in inclusive design can achieve expanded market reach and sustained competitive advantages, while those treating these considerations as overhead face compounding disadvantages as network effects and data advantages accrue to early movers.}


\begin{document}

\maketitle
\footnotetext{$^{\dagger}$These authors contributed equally to this work.}

\section{Introduction}
\label{sec:intro}

In 1934, the National Library Service introduced audiobook technology exclusively for individuals with visual impairments \citewithlink{loc2024history}{https://www.loc.gov/nls/who-we-are/history/}. Today, this assistive technology has transformed into an \$8.1 billion market projected to reach \$81 billion by 2034, with the majority of users having no disability whatsoever \citewithlink{emr2024audiobooks}{https://www.expertmarketresearch.com/reports/audiobooks-market} \citewithlink{ram2024audiobooks}{https://www.researchandmarkets.com/reports/6071241/audiobooks-global-strategic-business-report}. Mainstream consumers discovered that listening to books while commuting or exercising enhanced their productivity. Text-to-speech technology followed a similar trajectory: developed for users with visual impairments and reading disabilities, it now serves a broad population for multitasking and language learning. Only 5 percent of college students have learning disabilities requiring assistive technology, yet 26 percent voluntarily use text-to-speech tools \citewithlink{edtech2023tts}{https://www.edtechdigest.com/2023/01/13/text-to-speech-no-longer-just-for-students-with-disabilities/}. The market is expected to grow from \$3.6 billion in 2023 to \$14.6 billion by 2033 \citewithlink{gmi2024tts}{https://www.gminsights.com/industry-analysis/text-to-speech-market}.

This transformation illustrates a counterintuitive principle: when innovators design solutions to address the constraints of marginalized users, they often create innovations valuable for everyone. Addressing edge cases demands robust, flexible solutions that transcend conventional limitations.

We term this phenomenon the \textbf{\textit{inclusive innovation dividend}}---when solutions engineered for diverse, constrained use cases generate superior economic returns in broader markets. This concept represents an untapped opportunity in medical AI with direct implications for healthcare economics. While traditional analyses emphasize efficiency gains within existing markets, the inclusive innovation dividend highlights how prioritizing fairness—by removing systemic barriers, expanding access, and designing for underserved populations—can unlock entirely new revenue streams.

Healthcare AI development, however, remains anchored to a perceived trade-off between equitable performance and technical excellence. This misconception transforms fairness from strategic differentiator into regulatory burden, causing executives to overlook the economic insight that drove audiobooks from serving thousands to capturing billions. Historically, this view has been reinforced by structural constraints—single-site or demographically narrow datasets, idealized workflows, English-only documentation, and assumptions about stable devices and high-bandwidth environments. The industry's common development practices for homogeneous populations sacrifices not only equity but also enhanced generalization, expanded market reach, and revenue potential that inclusive design delivers.



In this article, we present evidence that fairness in medical AI generates measurable economic returns through multiple mechanisms: market expansion via geographic scalability and trust-based adoption (Section \ref{sec:market_expansion}), risk mitigation through reduced litigation exposure and regulatory penalties (Section \ref{sec:risk_mitigation}), performance improvements that lower technical debt and enhance generalizability (Section \ref{sec:performance_dividend}), and competitive advantages in talent retention and clinical adoption (Section \ref{sec:talent_advantages}). We synthesize these findings into an evaluation framework that integrates FDA, and HHS guidance with health technology assessment methodologies to help investors identify sustainable competitive advantages through fairness and generalizability metrics (Section \ref{sec:strategic_investment}). Our analysis reveals the business case rests not on ethical arguments or regulatory compliance alone, but on evidence that inclusive design can drive quantifiable economic value.
\begin{figure}[tb]
\centering
\includegraphics[page=1, trim={0cm 0 0cm 0}, clip, width=\textwidth]{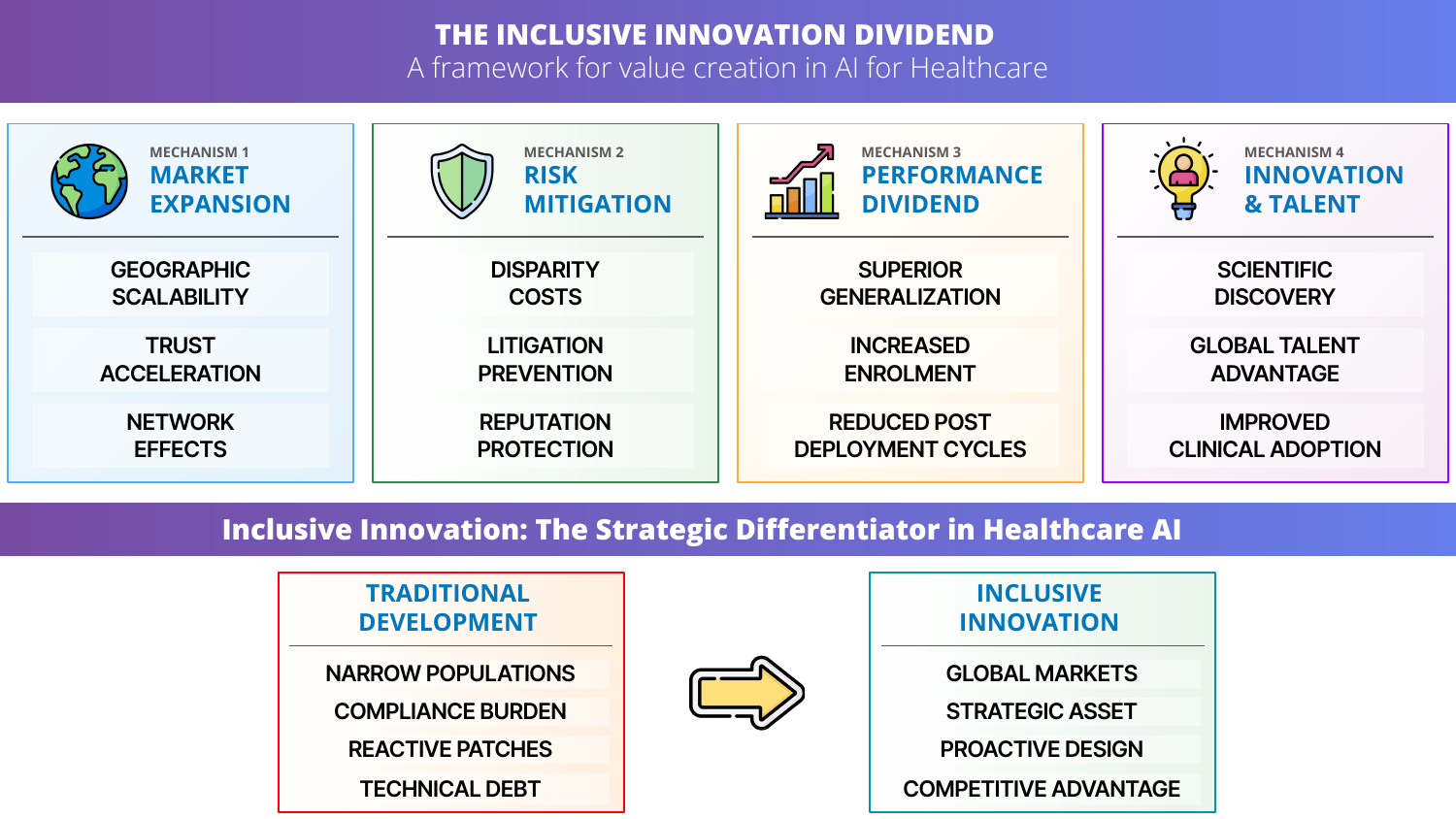}


\caption{The Inclusive Innovation Dividend in Healthcare AI. Four value-creation mechanisms demonstrate how solutions engineered for diverse, constrained use cases generate superior economic returns: market expansion through geographic scalability and trust acceleration; risk mitigation via disparity cost reduction and litigation prevention; performance dividends from superior generalization and reduced deployment cycles; and innovation advantages through scientific discovery and global talent attraction. The framework transforms fairness from compliance burden into strategic differentiator, paralleling how assistive technologies evolved from serving thousands to capturing billions.}
\label{fig:inclusive_innovation_dividend}
\end{figure}
\section{Market Expansion Through Algorithmic Fairness}
\label{sec:market_expansion}

Healthcare AI systems designed to perform equitably across diverse populations can expand markets instead of merely redistributing existing value. Investments in fairness create value through three mechanisms: (1) technical robustness that broadens the addressable market, (2) ``trust economics'' that accelerate adoption, and (3) network effects that amplify competitive advantage.

\subsection{Market-Creating Robustness and Geographic Scalability}

Healthcare AI models trained on homogeneous datasets often demonstrate performance degradation when deployed across diverse demographic contexts, creating commercial limitations beyond ethical concerns. The clinical AI fairness literature emphasizes that bias enters at multiple stages of the pipeline, requiring comprehensive mitigation strategies including pre-processing, in-processing, and post-processing approaches to ensure equitable performance \citewithlink{liu2025scoping}{https://www.nature.com/articles/s41746-025-01667-2} 

Medical imaging exemplifies how narrow training data constrains market potential. Traditional melanoma algorithms trained predominantly on lighter skin tones exhibit significantly higher misdiagnosis rates for darker-skinned patients. Recent research reveals these biases and proposes comprehensive skin-tone classification frameworks incorporating scales beyond traditional Fitzpatrick classifications \citewithlink{fitzpatrick1975soleil}{https://en.wikipedia.org/wiki/Fitzpatrick_scale}, including skin hue dimensions such as cold, neutral, warm, and olive undertones \citewithlink{montoya2025towards}{https://arxiv.org/abs/2411.12846}. Addressing these limitations represents not merely ethical advancement but fundamental market expansion, particularly in regions where darker skin tones predominate.

The geographic opportunity is substantial. Nearly 2 billion people globally lack adequate access to primary health services, with only 36 percent of the global nursing workforce serving rural areas despite these regions housing nearly half the world's population \citewithlink{balakrishnan2025artificial}{https://arxiv.org/abs/2508.11738}. This underscores the magnitude of untapped markets for AI tools capable of supporting care delivery in infrastructure-limited regions.

Successful deployments demonstrate economic viability --- telemedicine screening programs in rural India have achieved cost-effectiveness of \$1,320 per quality-adjusted life-year relative to no screening \citewithlink{ramasamy2021telemedicine}{https://pmc.ncbi.nlm.nih.gov/articles/PMC8725153                       /} and Google's partnership with local providers aims to deliver six million AI-supported retinal screenings in India and Thailand over ten years at no cost to patients \citewithlink{google2023ai-eye-care}{https://blog.google/around-the-globe/google-asia/arda-diabetic-retinopathy-india-thailand/}, illustrating how systems designed for diverse populations extend specialist care to previously inaccessible communities.

Constraint-driven innovation in diabetic retinopathy screening further highlights this potential. IDx-DR, designed specifically for primary care settings without ophthalmologist oversight, became the first FDA-approved autonomous AI diagnostic system \citewithlink{abramoff2018pivotal}{https://www.nature.com/articles/s41746-018-0040-6}, establishing regulatory precedent and creating new deployment models in non-specialist environments. Similarly, in Africa, deep learning-based AI has been developed to screen for referable and vision-threatening diabetic retinopathy using portable, non-mydriatic fundus cameras common in rural, low-resource settings \citewithlink{bellemo2019artificial}{https://www.thelancet.com/journals/landig/article/PIIS2589-7500(19)30004-4/fulltext}. These devices produce images with artifacts, variable lighting, and lower resolution compared to high-end Western equipment, yet models trained on such diverse, low-quality datasets achieve high accuracy, addressing the critical shortage of ophthalmologists and preventing blindness in underserved populations. Both approaches exemplify how inclusive designs for constrained contexts, whether specialist scarcity in primary care or equipment limitations in rural areas, not only ensure technical robustness but also unlock funding and scalability by enabling cross-border applicability and market expansion from low-income nations.

\subsection{Trust Economics and Market Acceleration}

Trust fundamentally determines healthcare AI adoption, with perceived fairness critically influencing user confidence. A 2025 JAMA Network Open study found 65.8 percent of US adults report low trust that health systems will use AI responsibly, while 57.7 percent have low trust that AI tools will not cause harm. Critically, discrimination experiences were significantly associated with reduced trust (OR 0.66; 95\% CI, 0.48–0.92) \citewithlink{nong2025patients}{https://jamanetwork.com/journals/jamanetworkopen/fullarticle/2830240}. These findings demonstrate that market penetration requires addressing equity concerns alongside technical performance.

Well-intentioned systems can inadvertently amplify biases, undermining trust. A Northwestern study of AI-assisted dermatology revealed that while AI assistance increased overall diagnostic accuracy, it disproportionately improved accuracy for lighter skin tones, thereby widening performance gaps \citewithlink{northwestern2024racial-bias}{https://news.northwestern.edu/stories/2024/02/new-study-suggests-racial-bias-exists-in-photo-based-diagnosis-despite-assistance-from-fair-ai/}. Such findings emphasize continuous fairness monitoring throughout deployment to maintain trust and ensure sustained adoption.

Organizations demonstrating fairness through comprehensive bias auditing, transparent subgroup performance reporting, and inclusive development processes position themselves to capture markets where trust deficits currently limit adoption. Reviews consistently identify fairness and transparency as critical enablers of healthcare worker confidence, while algorithmic opacity erodes clinical adoption \citewithlink{tun2025trust}{https://www.jmir.org/2025/1/e69678}.

\subsection{Network Effects and Ecosystem Value}

Inclusive AI systems generate positive externalities creating self-reinforcing improvement cycles. By serving diverse populations, these systems accumulate richer datasets that improve predictive accuracy for all users while creating competitive barriers. Economic analyses demonstrate that user contributions enhance predictions, attracting additional users and creating powerful data network effects \citewithlink{haftor2023pathway}{https://www.sciencedirect.com/science/article/pii/S0148296323006033}.

Following reverse innovation patterns, AI tools validated in constrained settings frequently discover unexpected applications in developed markets. The robustness required for environments with limited infrastructure and varied patient presentations often yields superior generalization capabilities. Comprehensive bias detection across the machine learning pipeline—including preprocessing for training imbalances, fairness-aware loss functions, and threshold calibration—ensures equitable deployment that strengthens platform value \citewithlink{chinta2025ai}{https://pmc.ncbi.nlm.nih.gov/articles/PMC12091740/ }.

Healthcare AI prioritizing fairness from inception creates expanding markets rather than competing for fixed share. Through enhanced robustness, trust-based adoption acceleration, and network effects, equitable design drives commercial growth while advancing health equity.
\section{Risk Mitigation and Regulatory Advantage}
\label{sec:risk_mitigation}

\subsection{Economic Burden of Algorithmic Bias}

Healthcare disparities represent substantial economic inefficiency. A 2018 U.S. study estimated the annual burden of racial and ethnic health disparities at \$451 billion, approximately 2\% of U.S. GDP \citewithlink{laveist2023economic}{https://jamanetwork.com/journals/jama/article-abstract/2804818}. When healthcare algorithms inadvertently perpetuate these disparities---as documented in Obermeyer et al.'s analysis of an algorithm affecting 200 million Americans that differentially reduced care access for Black patients \citewithlink{obermeyer2019dissecting}{https://www.science.org/doi/abs/10.1126/science.aax2342}---they may compound existing inefficiencies. Organizations developing equitable AI systems could potentially address these disparities while creating value.

\subsection{Economic Asymmetry of Prevention versus Remediation}

Available evidence suggests that proactive fairness implementation may be economically favorable compared to post-deployment remediation. Analysis of recent healthcare AI ventures indicates substantial financial and reputational risks associated with inadequate validation: IBM's divestiture of Watson Health to Francisco Partners involved reported losses exceeding \$3 billion \citewithlink{statibmwatson2022}{https://www.statnews.com/2022/01/21/ibm-watson-health-sale-equity/}, while Babylon Health's valuation declined from \$4.2 billion in its 2021 SPAC merger to bankruptcy within two years \citewithlink{babyloncollapse2023}{https://techcrunch.com/2023/08/31/the-fall-of-babylon-failed-tele-health-startup-once-valued-at-nearly-2b-goes-bankrupt-and-sold-for-parts/}. Notably, major academic medical centers including Memorial Sloan Kettering and MD Anderson terminated partnerships with IBM Watson following performance concerns, with MD Anderson spending \$62 million before cancellation \citewithlink{mdanderson2017}{https://www.medscape.com/viewarticle/876070}. These reputational impacts appear to persist beyond immediate financial losses, as evidenced by the absence of successful recovery attempts among healthcare AI ventures experiencing fundamental technology failures. In contrast, incremental costs for comprehensive bias testing and validation typically represent 16--18\% above baseline development costs---a difference measured in hundreds of thousands rather than billions.

\subsection{Legal and Regulatory Considerations}

Recent litigation highlights potential risks associated with algorithmic decision-making. In 2023, legal proceedings were initiated against Cigna regarding its PxDx algorithm's claim review processes, with allegations of insufficient physician oversight \citewithlink{cigna2023lawsuit}{https://www.statnews.com/2023/07/24/cigna-lawsuit-claim-denials/}. Similar litigation involving UnitedHealthcare's nH Predict algorithm raised questions about post-acute care determination accuracy \citewithlink{united2023lawsuit}{https://www.statnews.com/2023/11/14/unitedhealth-class-action-lawsuit-algorithm-medicare-advantage/}. While these cases remain in litigation, they illustrate both financial exposure and potential reputational consequences, as public scrutiny of algorithmic decision-making intensifies.

The regulatory environment continues to evolve. The 2024 HHS rules under Section 1557 now explicitly address ``patient care decision support tools'', requiring covered entities to implement measures to identify and address potential discriminatory impacts \citewithlink{mello_2024}{https://jamanetwork.com/journals/jama-health-forum/fullarticle/2823255}\citewithlink{federalregister_2024}{https://www.federalregister.gov/documents/2024/05/06/2024-08711/nondiscrimination-in-health-programs-and-activities}. Additionally, FDA's 2025 draft guidance emphasizes transparency and bias assessment in AI-enabled medical devices \citewithlink{fda2025guidance}{https://www.fda.gov/regulatory-information/search-fda-guidance-documents/artificial-intelligence-enabled-device-software-functions-lifecycle-management-and-marketing}.

Although monetary penalties for algorithmic bias have not yet been assessed in the U.S., regulatory activity suggests increasing attention to these issues. State-level initiatives include California Attorney General's August 2022 information requests to 30 hospital executives regarding algorithmic disparities \citewithlink{bonta2022investigation}{https://oag.ca.gov/news/press-releases/attorney-general-bonta-launches-inquiry-racial-and-ethnic-bias-healthcare} and Texas's September 2024 settlement with Pieces Technologies regarding AI performance claims, though the latter involved no monetary penalty \citewithlink{paxton2024pieces}{https://www.texasattorneygeneral.gov/news/releases/attorney-general-ken-paxton-reaches-settlement-first-its-kind-healthcare-generative-ai-investigation}.

Beyond risk mitigation, fairness-oriented development may facilitate market access. Models meeting stringent equity requirements could potentially enter markets with comprehensive AI regulations more readily. The EU AI Act's requirements for representative datasets and bias mitigation in high-risk clinical applications exemplify emerging global standards\citewithlink{van2023ai}{https://www.overleaf.com/project/6824d14fb8feb97bad734318}. Organizations developing to these specifications may find reduced barriers to international deployment. Systems with documented fairness assessments, representative training data, and appropriate oversight mechanisms may experience advantages including streamlined regulatory review, reduced post-market modification requirements, and sustained institutional partnerships\citewithlink{shabani2023fairness}{https://houstonhealthlaw.scholasticahq.com/api/v1/articles/128626-introducing-a-fairness-checkpoint-for-data-quality-and-evidence-during-regulatory-review-of-ai-ml-enabled-medical-devices.pdf}. These factors suggest that fairness investments could serve risk management, reputation preservation, and competitive positioning objectives.
\section{Performance Fairness Dividend}
\label{sec:performance_dividend}

A misconception in the development of medical AI is that fairness comes at the expense of performance. In practice, the opposite is often true: addressing bias can yield larger and more clinically meaningful gains than marginal improvements in already well-served populations. From a statistical and econometric perspective, many fairness issues reflect forms of sample selection bias, omitted variable bias, or distributional shift—factors that undermine a model’s external validity and real-world utility. Correcting these biases often improves out-of-sample performance, particularly in underrepresented groups where error rates are highest. Fairness interventions, as a result, frequently enhance accuracy, generalizability, and operational efficiency—outcomes that translate directly into business value.

\subsection{Fairness Interventions as Performance Upgrades}

The relationship between fairness and performance in medical AI merits careful examination, as evidence suggests that addressing algorithmic bias can yield substantial clinical improvements. Obermeyer and colleagues' analysis provides an illustrative case \citewithlink{obermeyer2019dissecting}{https://www.science.org/doi/abs/10.1126/science.aax2342}. Their evaluation of a care management algorithm affecting 200 million Americans identified systematic underestimation of Black patients' health needs resulting from the use of healthcare costs as a proxy for illness severity. This proxy variable reflected differential access patterns rather than clinical need. Recalibration using direct health indicators increased Black patient enrollment in high-risk care programs from 17.7\% to 46.5\% while maintaining or improving performance metrics for other demographic groups.

The Epic Sepsis Model offers a contrasting example of performance limitations in models deployed without comprehensive bias assessment. External validation across 38,455 hospitalizations yielded an area under the curve of 0.63, compared to vendor-reported values of 0.76--0.83, with 67\% of sepsis cases undetected and alerts generated for 18\% of all hospitalized patients \cite{wong2021epic}. The positive predictive value of 12\% resulted in substantial alert burden. Analysis identified dependence on variables such as prior antibiotic orders, which introduced temporal delays in alert generation \citewithlink{ross2021stat}{https://www.statnews.com/2021/09/27/epic-sepsis-algorithm-antibiotics-model/}. These findings suggest that systematic bias assessment may identify fundamental modeling limitations that affect overall performance.

\subsection{Clinical Accuracy Through Bias Correction}

Several mechanisms may explain the observed relationship between fairness-oriented development and improved clinical accuracy. 

First, constraining models to avoid proxy variables encourages identification of more direct clinical predictors. Healthcare utilization, influenced by socioeconomic factors and access barriers, correlates imperfectly with disease burden. Models utilizing direct health indicators demonstrate improved predictive validity across demographic groups.

Second, addressing measurement bias in clinical instruments enhances data quality. Pulse oximetry measurements exhibit systematic errors in patients with darker skin pigmentation, with documented three-fold higher rates of occult hypoxemia among Black patients when SpO$_2$ readings indicate 92-96\% saturation \citewithlink{sjoding2020racial}{https://www.nejm.org/doi/full/10.1056/NEJMc2029240}. Correction for such measurement bias has implications for all patients by reducing systematic error.

Third, fairness constraints may reduce shortcut learning, where models exploit spurious correlations specific to training distributions. These shortcuts often reflect overfitting to idiosyncrasies in the training data—essentially capturing noise rather than signal, or manifest as demographic proxies that exhibit poor generalization under distribution shift\citewithlink{brown2023detecting}{https://www.nature.com/articles/s41467-023-39902-7}. Bias mitigation techniques encourage learning of features with stronger causal relationships to outcomes\citewithlink{xiao2023fitness}{https://arxiv.org/abs/2305.14396}.

Fourth, demographic-aware calibration methods address subgroup-specific probability estimation errors that standard techniques often overlook. While standard calibration approaches optimize average performance across the full sample, they can obscure systematic miscalibration in minority or underrepresented groups\citewithlink{pleiss2017fairness}{https://arxiv.org/pdf/1709.02012}—effectively minimizing global loss at the expense of local reliability. By explicitly accounting for heterogeneity in predictive accuracy, demographic-aware methods improve both model fairness and statistical validity across subpopulations.  Fairness-constrained calibration methods have been shown to improve reliability across the prediction spectrum\citewithlink{shen2025exposing}{https://arxiv.org/pdf/2506.23298}.

\subsection{Technical Robustness and Sustainability}

Models developed with explicit fairness considerations demonstrate improved temporal stability, with performance maintenance observed across demographic shifts and evolving clinical practices\citewithlink{giguere2022fairness}{https://par.nsf.gov/servlets/purl/10334581}. This stability manifests in operational metrics including reduced false positive rates, decreased site-specific adjustment requirements, and lower post-deployment modification frequency. The interaction between these improvements suggests potential compound benefits, where initial fairness-driven enhancements facilitate subsequent optimizations\citewithlink{hasanzadeh2025bias}{https://www.nature.com/articles/s41746-025-01503-7}. These observations indicate that fairness considerations in model development may contribute substantially to the technical requirements for sustainable clinical deployment and the realization of anticipated healthcare improvements from artificial intelligence applications.
\section{Innovation and Talent Advantages}
\label{sec:talent_advantages}

Fairness and diversity serve as safeguards against groupthink, helping to expose failure modes at the margins—precisely the scenarios that challenge real-world deployments and can lead to unforeseen risks. Pluralistic development processes expand the search space of solutions and yield more resilient decision-making under uncertainty. In applied terms, diversity in teams and data increases the likelihood that learned representations are tethered to clinically relevant signal rather than spurious correlations that fail on subgroups, which accelerates generalizable discovery and decreases costly implementation rework. When the organizational objective is long-horizon competitiveness in regulated markets, diversity is best understood as a source of option value and a hedge against domain shift, not as a constraint on performance.

\subsection{Diverse Data and Teams Drive Discovery and Performance}

Large-scale, diverse cohorts systematically reveal biology invisible in homogeneous datasets, directly translating to new targets, better risk stratification, and broader market applicability. The All of Us Research Program's 2024 Nature report \cite{all2024genomic} released 245,388 clinical-grade whole genomes, with 77\% of participants from communities historically underrepresented in biomedical research and 46\% from underrepresented racial and ethnic minorities. The data include more than 1 billion genetic variants, with over 275 million previously unreported variants and more than 3.9 million with coding consequences. These figures underscore long-standing concerns that Eurocentric discovery pipelines miss population-specific signal, aligning with prior evidence that genome-wide analyses have been dominated by European-ancestry samples \cite{wojcik2019genetic}. 

For medical AI specifically, representative training distributions and evaluation cohorts are associated with less shortcut learning and improved out-of-distribution behavior—prerequisites for equitable clinical performance and efficient translational pathways \cite{heming2023benchmarking}. Investment in diverse cohorts therefore yields dual returns: increased clinical validity across subgroups and expanded viable indications and geographies that a product can address without costly recalibration.

Organizationally, plural leadership teams demonstrate superior financial outcomes. Companies in the top quartile for gender or ethnic diversity on executive teams are 39\% more likely to financially outperform peers \cite{dixon2023diversity}. Observed mechanisms include higher innovation revenue and faster problem solving rather than solely reputational effects \cite{hosny2022,galletta2022,andrevski2014racial}. These effects align with how cognitive variety improves problem solving and reduces correlated error in decision-making under uncertainty.

Across the clinical AI lifecycle, diverse teams better anticipate subgroup failure modes, align documentation with fairness-by-design expectations, and minimize rework during clinical integration. Plural perspectives improve dataset curation and labeling strategies, stress-test deployment assumptions across heterogeneous workflows, and increase the credibility of transparency artifacts such as model cards \cite{heming2023benchmarking,siddik2025datasheets}. The resulting performance dividend compounds from data collection through validation and post-market monitoring, yielding faster iteration on edge cases and fewer corrective releases.

\subsection{Talent Pipelines and Clinical Adoption}

Sustained AI leadership depends on attracting and retaining global talent. More than half of the U.S. AI-relevant workforce is foreign-born, and roughly two-thirds of graduate students in AI-related programs are international \cite{zwetsloot2019}. Among U.S. computer science Ph.D. recipients in 2024, 61\% earned their bachelor’s degree outside the United States. Of these, China accounted for 23\%, India for 10\%, and Iran for 4\% \cite{ncses2025}. Historically, about 80\% of U.S.-trained Ph.D. graduates stay in the country five years after graduation with particularly strong retention among graduates from Iran (95\%), Taiwan (93\%), China (91\%), and India (89\%) \cite{zwetsloot2019}. However, recent data from August 2025 indicate a 19\% decline in international student arrivals compared to the previous year, including decreases from China (12\%), India (44\%), and Iran (86\%) \cite{nytimes_decline}. These reductions are likely driven by sweeping travel bans, visa revocations, delayed visa processing, extreme vetting practices, and grant cancellations that create uncertainty and diminish the United States’ appeal to prospective talent \cite{nytimes_decline}.

Meanwhile, the global training pipeline is diversifying: 47\% of top AI researchers completed undergraduate education in China versus 18\% in the United States \cite{macropolo2024}. Beyond the United States, AI competitiveness increasingly depends on cross-border collaboration and inclusive research environments. European institutions competing for talent emphasize diverse, international teams as assets for regulatory navigation under frameworks like the EU AI Act, which mandates representative datasets and bias mitigation for high-risk clinical AI. Similarly, emerging AI hubs in Singapore, India, Brazil, and the Persian Gulf region position inclusivity as differentiators in global talent markets, recognizing that restrictive immigration policies or homogeneous work cultures create competitive disadvantages in attracting the multilingual, multicultural expertise required for deploying AI across diverse patient populations and regulatory jurisdictions. Organizations and ecosystems that lower barriers to entry and advancement for international talent better recruit and retain scarce expertise, sustaining competitive advantage in talent-constrained markets.

Clinician trust and workflow fit strongly influence whether models are used as intended or circumvented by shadow processes. Both improve when systems are transparent, auditable, and fair. Fairness-by-design practices that surface subgroup performance in model cards reduce uncertainty at the point of care and streamline site-specific calibration \cite{heming2023benchmarking,siddik2025datasheets}. In a national survey, 65.8\% of respondents reported low trust in their healthcare system to use AI responsibly, with prior discrimination experiences significantly reducing trust (OR, 0.66; 95\% CI, 0.48–0.92) \cite{nong2025patients}. Technical parity without equity will not secure durable acceptance. Embedding fairness in documentation and governance aligns with emerging safety expectations, reducing approval friction and supporting safer iteration cycles \cite{ratwani2024patient}.

The evidence supports a clear strategic thesis: diversity and fairness increase the probability of discovery, raise organizational performance, strengthen talent pipelines, and improve clinical adoption—advantages that compound into competitive positioning in regulated health markets. Organizations that treat inclusion as a first-order design constraint are better positioned to translate scientific breakthroughs into equitable products with fewer deployment surprises and more resilient market traction.

\section{Strategic Investment Framework}
\label{sec:strategic_investment}

\subsection{Translating Value Mechanisms into Investment Strategy}

The preceding sections identified four interconnected mechanisms through which inclusive innovation generates economic value in healthcare AI: market expansion via inclusive design, liability risk reduction, performance enhancement through bias mitigation, and competitive advantages from diverse innovation ecosystems. While these mechanisms provide conceptual justification for fairness investment, healthcare organizations require operational frameworks to evaluate and prioritize AI systems that can capture these benefits. 

Current evaluation methodologies emphasize technical accuracy metrics while underweighting factors critical to real-world deployment success—demographic generalizability, regulatory resilience, and sustainable market positioning. This evaluation gap becomes increasingly problematic as healthcare AI transitions from pilot projects to scaled clinical implementation. We propose the Healthcare AI Inclusive Innovation Framework (HAIIF) as a structured scoring system to quantify an AI system's potential across all four value-creation mechanisms, enabling systematic investment prioritization.

\subsection{The HAIIF Scoring Framework}

HAIIF evaluates healthcare AI systems across four interdependent domains, each mapped to established value mechanisms and weighted according to its contribution to sustainable deployment success. Table \ref{tab:haiif_scoring} presents the proposed scoring criteria with specific performance thresholds.

\begin{table}[hptb]
\centering
\caption{HAIIF Scoring Framework}
\label{tab:haiif_scoring}
\begin{tabular}{p{7cm}cc}
\toprule
\textbf{Domain / Indicator} & \textbf{Points} & \textbf{Threshold} \\
\midrule
\textbf{Fairness \& Equity (30\% weight)} & & \\
~~• Comprehensive bias audits & 25 & All protected classes \\
~~• Training data diversity & 25 & $\geq$30\% minorities \\
~~• Fairness metrics compliance & 25 & All metrics passed \\
~~• Demonstrated disparity reduction & 25 & Measurable improvement \\
\midrule
\textbf{Regulatory \& Trust (15\% weight)} & & \\
~~• FDA engagement & 10 & Formal engagement \\
~~• Clinical explainability & 30 & Interpretable outputs \\
~~• Continuous monitoring system & 30 & Real-time tracking \\
~~• Clinician adoption rates & 30 & $>$80\% acceptance \\
\midrule
\textbf{Generalizability \& Technical Robustness (35\% weight)} & & \\
~~• External validation performance maintenance & 25 & $\geq$90\% of original \\
~~• Multi-site validation coverage & 20 & $\geq$5 systems \\
~~• Performance stability over time & 25 & $\geq$2 years \\
~~• Cross-demographic variance & 30 & $<$10\% \\
\midrule
\textbf{Economic \& Innovation Value (20\% weight)} & & \\
~~• Market expansion potential & 30 & $>$25\% increase \\
~~• 5-year Total Cost of Ownership (TCO) analysis & 25 & Positive ROI \\
~~• Network effects evidence & 25 & Demonstrated value \\
~~• Research collaboration & 20 & Active partnerships \\
\bottomrule
\end{tabular}


\begin{tabular}{p{1.5cm}p{1.8cm}p{8.7cm}}
\textbf{Tier} & \textbf{Score Range} & \textbf{Investment Recommendation} \\
\midrule
\textbf{Tier 1} & 80--100 & Demonstrates inclusive innovation dividend with superior market performance and sustainable competitive moats; warrants maximum resource allocation. \\
\textbf{Tier 2} & 65--79 & Strong technical foundation requiring targeted fairness improvements; conditional funding recommended. \\
\textbf{Tier 3} & 50--64 & Early-stage innovation with breakthrough potential but significant development needs; requires careful oversight. \\
\textbf{Tier 4} & $<$50 & Fundamental design flaws requiring  reconceptualization before investment consideration. \\
\bottomrule
\end{tabular}
\end{table}

The domain weightings in Table \ref{tab:haiif_scoring} are calibrated for medical imaging applications, which represent 84.4\% of FDA-authorized AI/ML medical devices \citewithlink{singh2025ai}{https://www.nature.com/articles/s41746-025-01800-1}. Organizations may adjust weightings based on specific application characteristics—for instance, population health tools might increase the fairness domain weight, while acute care diagnostics might emphasize regulatory compliance. Each domain maps directly to value mechanisms: generalizability enables sustained performance across diverse settings, fairness metrics predict market expansion potential, regulatory readiness reduces deployment friction, and economic indicators capture network effects and competitive positioning.

Investment tiers guide resource allocation based on composite scores. Tier 1 systems (80-100 points) demonstrate clear pathways to capturing inclusive innovation dividends, justifying accelerated deployment with full organizational support. Tier 2 (65-79) represents technically sound systems where targeted fairness enhancements could unlock substantial additional value. Tier 3 (50-64) encompasses promising innovations requiring significant development investment before market readiness. Tier 4 ($<$50) indicates fundamental architectural or methodological limitations suggesting that resources would be better allocated elsewhere.

\subsection{Proposed Implementation Economics}

We propose that organizations budget 15-20\% above baseline development costs for comprehensive fairness implementation in healthcare AI systems. This incremental investment would be distributed across four critical areas: data diversity and representation infrastructure (3-5\% of baseline), fairness testing and validation frameworks (4-7\%), enhanced regulatory documentation and monitoring systems (4-6\%), and ongoing performance assessment across demographic groups (3-4\%). These percentages represent our framework recommendations based on the economic mechanisms identified in this analysis.

To illustrate application of these principles, Supplementary Table S1 presents a detailed cost breakdown for a hypothetical diabetic retinopathy screening system. In this example, an 18.3\% incremental investment above the \$3.44 million baseline enables comprehensive fairness implementation across all HAIIF domains. This hypothetical allocation demonstrates how targeted investments in inclusive design could potentially achieve Tier 1 classification while maintaining cost efficiency. The example is intended as guidance for resource planning rather than a prescriptive template, as actual costs will vary based on application complexity, target populations, and existing organizational capabilities.

\subsection{Implementation Pathway Recommendations}

Organizations adopting HAIIF should consider a phased implementation approach that builds capabilities systematically while demonstrating value at each stage:

\textbf{Phase 1: Assessment and Prioritization.} Establish baseline HAIIF scores for existing AI portfolio, identify systems with highest potential for improvement relative to investment, and implement quick wins through enhanced documentation and basic demographic reporting. This phase typically requires 2-3 months and minimal incremental investment while establishing measurement infrastructure.

\textbf{Phase 2: Systematic Integration.} Embed fairness considerations into development workflows, establish partnerships for diverse data collection, implement automated bias detection systems, and conduct multi-site validation studies. This 6-9 month phase represents the core transformation and requires the majority of incremental investment.

\textbf{Phase 3: Continuous Optimization.} Deploy real-time monitoring for demographic performance variations, establish feedback loops with clinical users across diverse settings, and iterate based on deployment evidence. This ongoing phase ensures sustained value capture and positions organizations to adapt as fairness standards evolve.

Strategic considerations for resource allocation include market characteristics, regulatory environment, and organizational capabilities. Applications targeting nationally diverse populations or high-stakes clinical decisions may justify investment toward the upper end of the 15-20\% range. Specialized tools for well-characterized populations might achieve adequate fairness with more modest investment. Organizations should avoid under-investment below 15\%, which risks inadequate infrastructure for meaningful improvement, while investment requirements exceeding 25\% may signal need for architectural redesign rather than incremental enhancement.

\subsection{Strategic Implications for Healthcare AI Development}

HAIIF provides a structured approach to portfolio management in healthcare AI, enabling organizations to systematically identify products warranting fairness investment versus those requiring fundamental reconceptualization. This portfolio-level perspective becomes increasingly valuable as regulatory frameworks evolve and market expectations for AI fairness crystallize. Organizations can optimize resource allocation across their development pipeline, accelerating high-potential systems while avoiding continued investment in fundamentally limited approaches.

The framework also facilitates stakeholder alignment across technical, clinical, and business teams. By quantifying fairness alongside traditional performance metrics, HAIIF creates common vocabulary for evaluating AI investments comprehensively. This shared understanding accelerates decision-making and reduces friction between teams with different priorities—for instance, helping clinical champions understand why certain technical limitations require addressing before deployment, or demonstrating to business leaders how fairness investments create market opportunities rather than merely adding costs.

Looking forward, we anticipate that frameworks like HAIIF will become standard practice as healthcare AI markets mature. Early adopters who systematically implement inclusive innovation principles will establish competitive advantages through regulatory precedents, talent attraction, and data network effects that compound over time. As academic medical centers, regulatory bodies, and payers increasingly recognize that technical performance alone insufficiently predicts real-world value, comprehensive evaluation frameworks will become prerequisites for market access rather than differentiators. Organizations that begin building these capabilities now will be best positioned to lead in an ecosystem where fairness and performance are understood as complementary rather than competing objectives.
\subsection{Limitations and Future Opportunities}
\label{sec:limitations}

Given the nascency of the field at the intersection of AI economics, healthcare delivery, and algorithmic fairness, several methodological considerations merit discussion. The proposed investment range of 15-20\% above baseline development costs derives from triangulation across industry reports, expert consultation, and documented FDA submission patterns, though peer-reviewed cost-effectiveness analyses specific to fairness interventions remain limited. Randomized controlled trials comparing development approaches are impractical in commercial settings; therefore, our framework synthesizes available evidence from peer-reviewed studies on algorithmic fairness, documented market outcomes, and regulatory guidance documents. Heterogeneity in fairness metrics and proprietary constraints on financial disclosure introduce uncertainty ranges consistent with early-stage health technology assessments. These methodological considerations parallel those in establishing economic frameworks for digital therapeutics and companion diagnostics, where preliminary estimates underwent iterative refinement as real-world evidence accumulated.

The evolving regulatory landscape presents both analytical challenges and research opportunities. FDA guidance documents and international frameworks including the EU AI Act continue to mature, potentially shifting the cost-benefit equilibrium presented here. Our HAIIF scoring thresholds and domain weightings represent initial benchmarks intended as a starting point for organizational assessment rather than absolute standards, requiring prospective validation across heterogeneous clinical applications and organizational contexts. These dynamics, however, create opportunities for rigorous health economic research. Prospective cohort studies can now be designed to quantify fairness return on investment systematically. Longitudinal analyses tracking economic outcomes across product lifecycles will enable empirical validation of the inclusive innovation dividend hypothesis. Multi-stakeholder collaborations could establish standardized metrics for fairness cost-effectiveness, analogous to quality-adjusted life year frameworks. As healthcare AI systems incorporating fairness-by-design principles achieve broader clinical deployment, the expanding evidence base will transform our theoretical framework into empirically-validated investment guidance, establishing fairness economics as a quantifiable discipline within digital health assessment.

\section{Conclusion}
\label{sec:conclusion}

The evidence presented resolves a false dichotomy: fairness in medical AI is not traded against performance or profitability—it enables both. Like audiobooks' transformation from an assistive technology to an \$81 billion industry, healthcare AI designed for constrained use cases creates mainstream value. Organizations implementing inclusive design capture four compounding returns: market expansion through geographic reach and trust acceleration, risk mitigation worth billions in avoided remediation costs, performance gains through reduced shortcut learning and superior generalization, and competitive advantages in talent acquisition and clinical adoption. The inclusive innovation dividend transforms fairness from compliance burden to strategic differentiator, with our analysis indicating 18\% incremental investment can yield 25-40\% market expansion.

Organizations implementing these principles secure competitive moats through network effects and data advantages that compound over time. Those treating fairness as regulatory overhead face escalating disadvantages: constrained market access, accumulating technical debt, and models that fail under distribution shift. The divergence accelerates as inclusive systems generate richer datasets, strengthen stakeholder trust, and establish regulatory precedents. Healthcare AI designed for everyone works better for everyone—and the institutions that recognize this now will define the industry's next decade.
\newpage
\section*{Supplementary Material}
\label{sec:supplementary}

\subsection*{Implementation Cost Analysis}

Table~\ref{tab:cba_improved} provides a detailed cost breakdown for implementing inclusive innovation in a representative healthcare AI system—diabetic retinopathy screening using 2D fundus imaging. The estimates are derived from interviews with engineering managers and technology executives at healthcare AI startups who have direct experience developing and deploying FDA-approved medical devices for 2D imaging applications with two or fewer clinical indications. This example illustrates practical resource allocation across the HAIIF framework domains.

The analysis demonstrates that comprehensive fairness implementation requires an 18.3\% incremental investment (\$770,000) above the \$3.44 million baseline for FDA-approved AI product. Notably, modern regulatory requirements already incorporate partial diversity considerations in the baseline, transforming inclusive innovation from a substantial new expense into a strategic upgrade that differentiates market leaders from merely compliance-focused competitors. The largest incremental investments support multi-site clinical validation in underserved populations (\$295,000) and development of fairness-aware algorithms (\$315,000), both of which directly contribute to market expansion potential and performance improvements discussed in the main text.

This breakdown serves as guidance for budget planning and can be scaled proportionally for applications of different complexity. Organizations could adjust allocations based on their specific technical requirements, target populations, and existing infrastructure.

\begin{table}[htbp]
\centering
\caption{Healthcare AI Development Costs: Realistic Baseline with Inclusive Innovation Increment\\
\small\textit{Based on 2D medical imaging AI device (e.g., diabetic retinopathy screening)}}
\label{tab:cba_improved}

\begin{tabular}{p{8.5cm} r r}
\toprule
\textbf{Activity} & \textbf{Baseline} & \textbf{Additional Cost} \\
& & \textbf{(Inclusive Innovation)} \\
\midrule

\rowcolor{gray!10}
\multicolumn{3}{l}{\textbf{DATA \& VALIDATION INVESTMENTS}} \\
Diverse fundus image collection & \$150,000 & \$25,000 \\
\quad\textit{Base: FDA-required diversity | Add: Global populations} & & \\
Multi-geographic dataset acquisition & \$75,000 & \$125,000 \\
\quad\textit{Base: US regions | Add: Africa, Asia datasets} & & \\
Bias-aware data preprocessing and quality assessment & \$35,000 & \$15,000 \\
\quad\textit{Base: Standard QA | Add: Bias-specific validation} & & \\
\textbf{Subtotal} & \textbf{\$260,000} & \textbf{\$165,000} \\
\addlinespace

\rowcolor{gray!10}
\multicolumn{3}{l}{\textbf{MULTI-SITE CLINICAL VALIDATION}} \\
Clinical trial across diverse populations (10+ sites) & \$850,000 & \$150,000 \\
\quad\textit{Base: FDA diversity req | Add: Under-served populations} & & \\
Community health center validation partnerships & \$100,000 & \$125,000 \\
\quad\textit{Base: Some diverse sites | Add: Full rural/urban coverage} & & \\
Reference standard analysis by demographic subgroups & \$200,000 & \$0 \\
\quad\textit{Base: FDA requires subgroup analysis | Add: Already included} & & \\
Statistical validation of bias mitigation effectiveness & \$150,000 & \$20,000 \\
\quad\textit{Base: Standard validation | Add: Fairness metrics testing} & & \\
\textbf{Subtotal} & \textbf{\$1,300,000} & \textbf{\$295,000} \\
\addlinespace

\rowcolor{gray!10}
\multicolumn{3}{l}{\textbf{FAIRNESS INFRASTRUCTURE (R\&D and Post-deployment)}} \\
Automated demographic bias detection systems & \$40,000 & \$85,000 \\
\quad\textit{Base: Performance monitoring | Add: Demographic-specific alerts} & & \\
Real-time fairness metrics monitoring and reporting & \$30,000 & \$70,000 \\
\quad\textit{Base: Standard metrics | Add: Fairness dashboards} & & \\
Cross-demographic performance tracking dashboards & \$80,000 & \$50,000 \\
\quad\textit{Base: Clinical dashboards | Add: Equity visualizations} & & \\
Continuous bias drift detection and alerting & \$20,000 & \$30,000 \\
\quad\textit{Base: Model drift alerts | Add: Demographic drift tracking} & & \\
\textbf{Subtotal} & \textbf{\$170,000} & \textbf{\$235,000} \\
\addlinespace

\rowcolor{gray!10}
\multicolumn{3}{l}{\textbf{AI ALGORITHM DEVELOPMENT}} \\
Training and Tuning (base requirement) & \$250,000 & \$0 \\
Fairness-constrained deep learning architecture & \$550,000 & \$175,000 \\
\quad\textit{Base: Robust architecture | Add: Fairness constraints} & & \\
Multi-demographic training data rebalancing & \$120,000 & \$80,000 \\
\quad\textit{Base: Class balancing | Add: Demographic balancing} & & \\
Adversarial debiasing technique implementation & \$60,000 & \$40,000 \\
\quad\textit{Base: Robustness testing | Add: Debiasing algorithms} & & \\
Explainable AI features for clinical decision support & \$80,000 & \$20,000 \\
\quad\textit{Base: FDA explainability | Add: Confounder Checks} & & \\
\textbf{Subtotal} & \textbf{\$1,060,000} & \textbf{\$315,000} \\
\addlinespace

\rowcolor{gray!10}
\multicolumn{3}{l}{\textbf{REGULATORY \& TRUST}} \\
Enhanced FDA and HTA documentation with fairness evidence & \$340,000 & \$10,000 \\
\quad\textit{Base: Standard FDA submission | Add: Fairness evidence} & & \\
Comprehensive demographic performance reporting & \$25,000 & \$15,000 \\
\quad\textit{Base: Required reporting | Add: Comprehensive equity data} & & \\
Transparency reporting and discrimination risk assessment & \$10,000 & \$10,000 \\
\quad\textit{Base: Risk assessment | Add: Discrimination-specific audit} & & \\
\textbf{Subtotal} & \textbf{\$375,000} & \textbf{\$35,000} \\
\addlinespace

\rowcolor{gray!10}
\multicolumn{3}{l}{\textbf{DEPLOYMENT \& INFRASTRUCTURE}} \\
Cloud infrastructure with bias monitoring capabilities & \$260,000 & \$15,000 \\
\quad\textit{Base: Standard monitoring | Add: Bias-specific features} & & \\
EHR integration with fairness reporting features & \$15,000 & \$10,000 \\
\quad\textit{Base: Clinical reporting | Add: Equity metrics} & & \\
\textbf{Subtotal} & \textbf{\$275,000} & \textbf{\$25,000} \\

\midrule
\midrule
\rowcolor{yellow!20}
\textbf{TOTAL PROJECT INVESTMENT} & \textbf{\$3,440,000} & \textbf{\$770,000} \\
\rowcolor{green!20}
\textbf{COMBINED TOTAL} & \multicolumn{2}{c}{\textbf{\$4,210,000}} \\
\rowcolor{blue!20}
\textbf{PERCENTAGE BREAKDOWN} & \textbf{81.7\%} & \textbf{18.3\%} \\
\midrule
\midrule

\multicolumn{3}{p{14cm}}{\textbf{Key Insights:}} \\
\multicolumn{3}{p{14cm}}{• \textbf{Realistic Baseline:} Modern FDA-approved AI already costs \$3.44M with partial inclusion} \\
\multicolumn{3}{p{14cm}}{• \textbf{True Incremental Cost:} Full inclusive innovation adds only \textbf{18.3\%} } \\
\multicolumn{3}{p{14cm}}{• \textbf{Strategic ROI:} 18\% investment → 25-40\% market expansion + reduced litigation risk} \\

\bottomrule
\end{tabular}
\end{table}
\newpage


\bibliographystyle{sn-nature}
\bibliography{references}

\newpage
\bmhead{Acknowledgements}
LAC is funded by the National Institute of Health through DS-I Africa U54 TW012043-01 and Bridge2AI OT2OD032701, the National Science Foundation through ITEST \#2148451, and a grant of the Korea Health Technology R\&D Project through the Korea Health Industry Development Institute (KHIDI), funded by the Ministry of Health \& Welfare, Republic of Korea (grant number: RS-2024-00403047).

\section*{Declarations}
\begin{itemize}
\item \textbf{Funding:} LAC is funded by NIH (DS-I Africa U54 TW012043-01, Bridge2AI OT2OD032701), NSF (ITEST \#2148451), and KHIDI (RS-2024-00403047). All other authors declare no funding.

\item \textbf{Conflict of interest/Competing interests:} The authors declare no competing interests.

\item \textbf{Ethics approval:} Not applicable.

\item \textbf{Consent to participate:} Not applicable.

\item \textbf{Consent for publication:} Not applicable.

\item \textbf{Data availability:} Not applicable.

\item \textbf{Code availability:} Not applicable.

\item \textbf{Author contributions:} 
\begin{itemize}
\item \textbf{Conceptualization and Design:} BU, AGA, AA, RK, LAC
\item \textbf{Market Analysis and Economic Framework:} BU, SP, SF
\item \textbf{Risk and Legal Analysis:} AA, BU, AP
\item \textbf{Performance and Technical Analysis:} AA, LG, RG
\item \textbf{Innovation and Talent Analysis:} AGA, AA
\item \textbf{Strategic Investment Framework:} SP
\item \textbf{Case Studies and Examples:} SP (DR example), HCK (Non-US examples)
\item \textbf{Writing - Original Draft:} BU, AGA, AA, SP, RK
\item \textbf{Writing - Review \& Editing:} All authors
\item \textbf{Visualization:} BU
\item \textbf{Supervision:} LAC
\end{itemize}
\end{itemize}

\end{document}